\documentclass[conference]{IEEEtran}
\IEEEoverridecommandlockouts
\usepackage{cite}
\usepackage{amsmath,amssymb,amsfonts}
\usepackage{algorithmic}
\usepackage{algorithm}
\usepackage{graphicx}
\usepackage{textcomp}
\usepackage{xcolor}
\usepackage{multicol}
\usepackage{tabularx}
\usepackage{CJKutf8}
\usepackage{booktabs} 
\usepackage{multirow} 
\usepackage{url}

\def\BibTeX{{\rm B\kern-.05em{\sc i\kern-.025em b}\kern-.08em
    T\kern-.1667em\lower.7ex\hbox{E}\kern-.125emX}}
\begin{document}

\title{Enhancing the Traditional Chinese Medicine Capabilities of Large Language
Model through Reinforcement Learning from AI Feedback
}

\author{
\IEEEauthorblockN{Song Yu\textsuperscript{1}, Xiaofei Xu\textsuperscript{2}, Fangfei Xu\textsuperscript{3}, Li Li\textsuperscript{1*}}
\IEEEauthorblockA{
\textsuperscript{1}\textit{School of Computer and Information Science, Southwest University, Chongqing, China}\\
lily@swu.edu.cn\\
\textsuperscript{2}\textit{School of Computing Technologies, RMIT University, Melbourne, Victoria 3000 Australia}\\
\textsuperscript{3}\textit{Donggexinli Station, Yongdingmenwai Community Health Service Center, Dongcheng, Beijing}\\
}
}

\maketitle

\begingroup\renewcommand\thefootnote{\textsuperscript{*}}
\footnotetext{Corresponding author.}
\endgroup
\begin{abstract}
Although large language models perform well in understanding and responding to user intent, their performance in specialized domains such as Traditional Chinese Medicine (TCM) remains limited due to lack of expertise. In addition, high-quality data related to TCM is scarce and difficult to obtain, making large language models ineffective in handling TCM tasks. In this work, we propose a framework to improve the performance of large language models for TCM tasks using only a small amount of data. First, we use medical case data for supervised fine-tuning of the large model, making it initially capable of performing TCM tasks. Subsequently, we further optimize the model's performance using reinforcement learning from AI feedback (RLAIF) to align it with the preference data. The ablation study also demonstrated the performance gain is attributed to both supervised fine-tuning and the direct policy optimization. The experimental results show that the model trained with a small amount of data achieves a significant performance improvement on a representative TCM task.
\end{abstract}

\begin{table*}[htbp]
\caption{Task Type}
\renewcommand{\arraystretch}{1.2}
\setlength{\tabcolsep}{2pt}
\centering
\begin{tabularx}{\textwidth}{|X|X|}
\hline
\multicolumn{1}{|c|}{\textbf{Task1: First visit }} & \multicolumn{1}{c|}{\textbf{Task2: Follow-up visit }} \\
\hline
\textbf{User input:}\par
\begin{CJK*}{UTF8}{gbsn} 
\textbf{患者}:窦某，女，64岁，2012年12月25日就诊。\par
\textbf{主诉}：入睡困难1月余。\par
\textbf{现病史}：患者1月前无明显诱因出现入睡困难，眠浅易醒，身上有力，心烦急躁，纳呆，口干，欲饮热水，二便调。
\par\par
\end{CJK*}
\textbf{Model output:}
\begin{CJK*}{UTF8}{gbsn}\par
\textbf{中医诊断}：不寐。\par
\textbf{治法}：清利肝胆，健脾养心。\par
\textbf{方剂}：小柴胡汤和归脾汤加减。\par
\textbf{处方}：柴胡10g，黄芩12g，半夏9g，炒白术15g，茯苓12g ，炒枣仁15g，龙眼肉15g，麦冬15g，合欢花15g，黄连6g，珍珠母30g，7付，水煎服7剂，水煎服。\end{CJK*}
&\textbf{User input:}\par
\begin{CJK*}{UTF8}{gbsn} 
\textbf{患者}:窦某，女，64岁，2012年12月25日就诊。\par
\textbf{主诉}：入睡困难1月余。\par
\textbf{现病史}：患者1月前无明显诱因出现入睡困难，眠浅易醒，身上有力，心烦急躁，纳呆，口干，欲饮热水，二便调。\par
\textbf{中医诊断}：不寐。\par
\textbf{治法}：清利肝胆，健脾养心。\par
\textbf{方剂}：小柴胡汤和归脾汤加减。\par
\textbf{处方}：柴胡10g，黄芩12g，半夏9g，炒白术15g，茯苓12g ，炒枣仁15g，龙眼肉15g，麦冬15g，合欢花15g，黄连6g，珍珠母30g，7付，水煎服7剂，水煎服。\par
\textbf{二诊}（12月31日）：服上方后睡眠改善不明显，入睡困难，眠浅易醒，心烦急躁，纳呆，口淡无味，口干欲饮，二便调，舌红，苔薄白，脉数。\par
\end{CJK*}
\textbf{Model output:}
\begin{CJK*}{UTF8}{gbsn}\par
\textbf{复诊开方}：熟地15g，炒白芍15g，枸杞15g，炒白术15g，炒枣仁15g，元肉15g，黄柏10g，淫羊藿10g，生龙骨30g，生牡蛎30g，夜交藤30g，7付，水煎服。
\end{CJK*}\\
\hline
\end{tabularx}
\end{table*}
\section{Introduction}
Language modeling, as an important approach to language understanding and generation, has been extensively studied over the past two decades, evolving from statistical language models to neural language models. Recently, pre-trained language models drown a lot of attention, and by pre-training Transformer models on large-scale corpora, these models have demonstrated their power in solving various natural language processing tasks\cite{plm}. Interestingly, when the parameter size exceeds a certain level, these language models not only improve performance, but also show capabilities that are not available in small-scale models. In general, large language models (LLMs) are Transformer-based language models containing billions or more parameters trained on large amounts of textual data, such as OpenAI's ChatGPT\cite{chatgpt} and GPT-4\cite{gpt4}, which are capable of understanding and answering a wide range of questions, and their performance on certain tasks even meets or exceeds that of humans. In addition, the open source community has rapidly introduced a series of LLMs, such as ChatGLM\cite{glm}, LLaMA\cite{llama}, Qwen\cite{qwen}, which have demonstrated impressive performance.

However, the training data for LLMs mainly come from the Internet and books and are mostly common-sense data, thus limiting them in domains such as Traditional Chinese Medicine (TCM). Additionally, high quality TCM data is scarce, and existing large language models perform poorly on TCM tasks. Despite the challenges, TCM LLM has great potential to provide value in answering questions, assisting doctors in diagnosis and prescribing.

In the medical domain, several medical language models have been proposed, such as Med-Palm\cite{med-palm}, DoctorGLM\cite{doctorglm}, etc., which show great promise in a variety of medical applications, such as medical question and answer, dialog systems, and text generation. However, almost all of these works focus on modern medicine, with very few addressing TCM. Developing a large model for the TCM domain is challenging. First, most TCM works are written in ancient Chinese, and many terms do not have corresponding explanations in modern Chinese, with large grammatical differences\cite{lozano2014basic}. Additionally, many theories and diagnostic and therapeutic methods in TCM lack uniform quantitative and objective standards and cannot be easily verified. Finally, TCM is not a widely applied discipline, and there is less information on related works, and high-quality data are even more difficult to obtain \cite{matos2021understanding}.

To address these limitations, we propose a framework for enhancing the performance of large-model TCM tasks with only a small amount of data. Specifically, we focus on two types of tasks: initial visit and follow-up visit, as shown in Table 1. Our framework consists of three stages. First, we collect a corpus of real medical cases and perform supervised fine-tuning on a open-source large language model, this step will steer the LLM into solving TCM tasks. Second, for each input, we instruct the model to generate multiple outputs to build a preference dataset. Considering the inefficiency and high cost of manual annotation, we introduced a reinforcement learning method based on AI feedback (RLAIF) to train language models using AI-generated feedback instead of human feedback. Finally, we use preference data to instruct the model's learning, enabling it to generate outputs that better align with user expectations.

The main contributions of this paper are as follows:
\begin{itemize}
    \item We develop a framework for enhancing the performance of TCM tasks using only a small amount of data, enabling the full process of training from supervised fine-tuning to RLAIF.
    \item We develop an automated labeling and ranking method using generative AI to build high-quality preference datasets instead of manual labor.
    \item We conduct several experiments on test datasets to demonstrate the effectiveness of our proposed framework.
\end{itemize}
\begin{figure*}[h]
\includegraphics[scale=1.0]{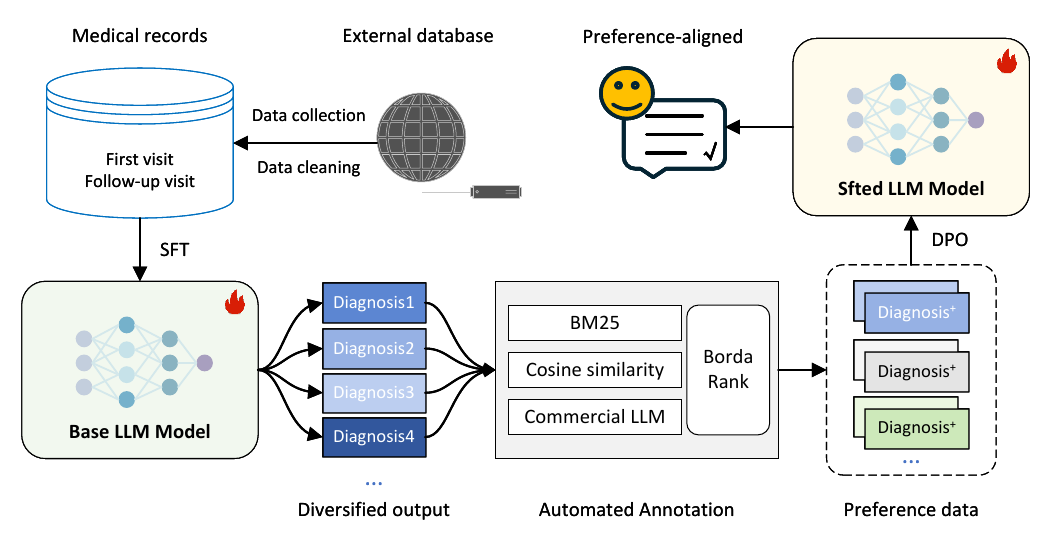}
\centering
\caption{The structure of our proposed framework includes data collection, supervised fine-tuning, automatic labeling, and direct preference optimization. After supervised fine-tuning, the model will generate multiple outputs, which are labeled using automatic labeling to obtain preference data. The model is further optimized using dpo and preference data.} \label{fig1}
\end{figure*}
\section{Related Work}
\subsection{Large language models}
In 2022, ChatGPT went live and attracted a lot of attention, triggering a heated discussion about LLM across the community. With the release of GPT-4 (2023) and GPT-4o (2024), the multimodal capability of large models was further improved, which set off a new wave of AI, however, OpenAI did not announce its training strategies and weights due to various factors. As a result, open source LLMs such as LLaMA and BLOOM\cite{bloom} quickly attracted a lot of attention from the research community once they were released. These models made public their training methods and weighting files, enabling researchers and developers to further train and improve the models on this basis. A large number of open source LLMs have also emerged in China, such as ChatGLM, DeepSeek\cite{deepseek}, Qwen, Baichuan\cite{baichuan}, and ERNIE\cite{ernie}. These open-source LLMs use rich Chinese corpus for training and optimization.

With the proliferation of these models, various optimization techniques have been explored to enhance their performance. Optimizing the performance of large language models involves various techniques, with RLHF (reinforcement learning from human feedback)\cite{rlhf} and RLAIF (reinforcement learning from AI feedback)\cite{rlaif} being two prominent methods. In RLHF, a reward model is trained to learn alignment based on human feedback. Once fine-tuned, this reward model can evaluate different outputs, scoring them according to the alignment preferences specified by humans. This feedback is then used to further refine the original language model. On the other hand, RLAIF involves directly linking a pre-trained, well-aligned model to the language model, allowing it to learn from larger and more aligned models. Research shows that RLAIF performs as well as, or even better than, human feedback (RLHF) in tasks such as text summarization, helpful dialogue generation, and harmless dialogue generation.

In a recent study known as Direct Preference Optimization (DPO)\cite{dpo} highlighted the complexity and instability of RLHF. They proposed an alternative approach by leveraging a mapping between reward functions and optimal policies. This mapping allows the constrained reward maximization problem to be optimized precisely through a single stage of policy training, effectively transforming it into a direct objective optimization based on human preference data. Their algorithm, termed DPO, is noted for its stability, performance, and computational efficiency, eliminating the need for fitting a reward model. They found that DPO surpasses RLHF in controlling sentiment generation and enhancing response quality in summarization.

\subsection{Large Language Models in Medical}
Currently, some progress has been made in the research of large language models in medical domain, but there are still some limitations and challenges. Due to the special characteristics of Traditional Chinese medicine, the medical domain tuned large language models released worldwide mainly focus on Western medicine, and most of them are in English as the main language, such as Google Med-PaLM series which has some limitations on the discovery and application of knowledge of TCM, and it is difficult to meet the special needs of TCM.

In China, research teams have begun to emphasize the development of Chinese medical LLMs, such as Huatuo GPT\cite{huatuogpt}, zhongjing\cite{yang2024zhongjing}, shennong-TCM\cite{zhu2023ChatMed} and DoctorGLM. However, the number of large models for TCM is relatively small compared with those for Western medicine. Most of the current so-called large models for Chinese medicine are not purely focused on traditional Chinese medicine. Instead, they are hybrid models that mix knowledge from Western medicine, Chinese medicine, and other related fields. These efforts may overly pursue breadth at the expense of depth. In addition, the quality of the data used to train those existing Chinese medicine LLMs rely varies, which affects the final results of the models. For example, Shennong used the TCM dataset generated by ChatGPT, but the quality cannot be guaranteed, and most of the tasks are common sense questions and answers rather than prescription tasks; Zhongjing collected a large amount of real-world data, but it was also not focused on TCM prescription tasks.

Despite the great potential of Large Language Models (LLMs) in healthcare, there are still some important and specific challenges that need to be addressed. When models are used for general knowledge quizzes, the impact of errors is not fatal to humans; however, this is not the case in the medical domain, where incorrect interpretations and answers can have serious consequences for patients. The accuracy and reliability of the information provided by LLMs can be life threatening, as it may affect medical decisions, diagnosis, and treatment plans. In addition, the definition of responsibility after using a LLM to aid in diagnosis is an issue that needs to be considered. Therefore, we need to ensure the quality of model output as much as possible.
\begin{table*}[h!]
\centering
\caption{Prompt example}
\label{tab:prompt_example}
\begin{tabular}{|m{15cm}|}
\hline
\textbf{Prompt Example} \\ \hline
Instruction: You are an intelligent assistant that specializes in solving medical-related problems for users. Please give specific Chinese medicine diagnosis, treatment and prescription based on the medical case provided to you by the user. \\You need to give a TCM diagnosis, treatment and prescription according to the symptoms provided to you by the user. \par I will give you an example: \par Patient: Ji, male, 43 years old. Initial diagnosis: 08/16/2021. \par Complaint: recurrent cough for 1 year. \par Medical history: the patient has had recurrent cough for the past 1 year. \par Physical examination: tongue red and dark. \par You need to give: \par 
TCM diagnosis: [specific diagnosis]. \par Treatment: [specific treatment]. \par Prescription: [name of prescription]. \par Prescription: [specific prescription]. \par Please note that some medical cases include follow-up visit data, and you will need to synthesize all the information to determine the next step in medication. Now, please give the TCM diagnosis, treatment and prescription based on the symptoms provided to your patient: \\ \hline
\end{tabular}
\end{table*}

\section{Method}
This section describes the process of building the framework, which is divided into data construction, supervised fine-tuning, reinforcement learning from AI feedback. Each step is discussed sequentially to reflect the research workflow. The integrated methodology flowchart is shown in Figure 1.
\subsection{Data Construction}
One of the challenges in training high-performance LLM models for TCM lies in obtaining high-quality data. A high quality corpus can greatly improve the performance of LLMs and even break the scaling law to some extent\cite{textbook}. The model needs not only theoretical data from TCM textbooks, but also professional data from real doctor-patient scenarios, which can reflect the specific conditions of patients and guide the addition, subtraction and proportion of medicines. 
In order to ensure the diversity of the medical corpus, we collect a variety of real medical text data from multiple sources, including open source data, proprietary data, and real medical consultations. These data cover most of the domains and symptoms of TCM, providing rich and detailed medical knowledge for the model.

In this paper, we will focus on the classical types of prescription tasks. Specifically, there are two types of prescription tasks: first visit and follow-up visit. The first visit task needs to issue a prescription based on the user's symptoms and examination results, while the follow-up visit task needs to synthesize all the previous follow-ups, prescriptions, and feedbacks from the user's medication to give subsequent medication suggestions.

\subsection{Supervised fine-tuning}
Supervised fine-tuning (SFT) is a key stage in empowering the model with dialog capabilities. With the help of high-quality doctor-patient dialog data, the model can effectively invoke the medical knowledge accumulated during the pre-training process to understand and answer the user's query. The goal of SFT is to teach the model how to understand and generate appropriate replies based on TCM diagnostic principles and treatment methods.

The SFT process consists of optimizing the model parameters to minimize the cross-entropy loss between the predicted output and the actual output for a given input set. In the SFT process, the model learns to generate prescriptions based on the initial diagnosis and subsequent diagnoses. This process involves providing the model with detailed case information including symptoms, diagnoses, and patient feedback to guide its learning process.

The cross-entropy loss used for training the model is defined as follows:

\begin{equation}
L = -\frac{1}{N} \sum_{i=1}^{N} \sum_{t=1}^{T} y_{it} \log(\hat{y}_{it})
\end{equation}

where \( N \) is the total number of samples, \( T \) is the sequence length, \( y_{it} \) is the true token at position \( t \) for the \( i \)-th sample, and \( \hat{y}_{it} \) is the predicted probability of the true token at position \( t \) for the \( i \)-th sample. This loss function measures the difference between the predicted token distribution and the actual token distribution, and the goal of training is to minimize this loss. The pseudo-code for this phase is shown in Algorithm 1.

Note a good prompt helps to stimulate the model's capability, so we designed a prompt to guide the model to generate the specified response, as shown in Table 2. We decide whether to use English or Chinese prompts based on the corpus used in the pre-training stage of the base model, but the output is uniformly specified to be in Chinese.
\begin{algorithm}[!ht]
    \renewcommand{\algorithmicrequire}{\textbf{Input:}}
	\renewcommand{\algorithmicensure}{\textbf{Output:}}
	\caption{Supervised Fine-tuning}
    \label{power}
    \begin{algorithmic}[1] 
        \REQUIRE  Initial policy $\pi_{\theta}$, Training dataset $D$, Number of epochs $N$, Batchsize $B$, Learning rate $\eta\in[0,1]$ 
	    \ENSURE SFTed policy $\hat{\pi}_\theta$ 
        
        \FOR {epoch=1,2,3,...,$N$}
            \FOR{$D_{batch}\sim D$}
                \STATE $(x,z)\sim D_{batch}$
                \STATE $\hat{z}\sim \pi_{\theta}(x)$
                \STATE $L = \text{CrossEntropyLoss}(\hat{z}, z)$
                \STATE $\theta \leftarrow \theta - \eta \nabla_{\theta} L$ //Update the parameters using any gradient descent algorithm
            \ENDFOR
        \ENDFOR
        
    \end{algorithmic}
\end{algorithm}

\subsection{Reinforcement Learning from AI Feedback}
Despite SFT accumulating medical knowledge and guiding conversational abilities, the model can still produce inaccurate, harmful, or unfriendly responses, which can have serious consequences in medical dialogues. We use RLAIF to improve the conversation process. Specifically, for each question, we guide the supervised fine-tuned model to generate diverse outputs, score them using a specific annotation model, and then rank them using Borda rank. Finally, we align the training using the DPO algorithm.
\subsubsection{AI Feedback for TCM}
Considering the specificity of medical conversations, we obtain feedback in three ways, make metrics from three different aspects of the output, and develop a sorted annotation rule.
\begin{itemize}
    \item \textbf{Lexical overlap:} This refers to cases where both the ground truth and the model output contain the same or similar terms. We use BM25, a well-established ranking function used by search engines \cite{robertson2009probabilistic}. The advantage of BM25 lies in its ability to weigh query terms according to their TF-IDF importance, thus providing a reliable measure of term overlap.
    \item \textbf{Semantic overlap:} This refers to cases where the ground truth and the model output contain semantically related content. Even if the model output does not contain the exact terms from the ground truth, they may still be relevant, such as "yin-yang imbalance" and "yang deficiency." We use the dense encoder model bert-base-chinese, which works by encoding ground truth and model output as dense vectors in a shared embedding space, respectively \cite{sun2021chinesebert}. It focuses on deeper semantic connections between words and phrases. The dot product of the two vectors then gives a measure of semantic similarity.
    \item \textbf{Model annotation:} The above two similarity measurement methods may be biased in TCM. Therefore, we also use DeepSeek-V2 and GPT-4o as annotation models, guiding them to evaluate the generated dialogues from the perspectives of completeness, professionalism, and fluency. Annotation questions come from the test set, and we provide both the questions and the standard answers to the annotation models. The annotation models will score the generated dialogues on a scale from 0 to 100.
\end{itemize}

To mitigate the differences caused by the scoring distribution of different models, ranking is used as a means of introducing regularization supervision signals. Specifically, when multiple rankings of candidate outputs are provided, we use a voting method, Borda Rank, to merge them into a unified ranking. For example, given $N$ items and $M$ individual rankings, the Borda ranking score $B_i$ of item $i$ is calculated as follow:
\begin{equation}
    B_i = \Sigma^M_{j=1}(N-rank_j(i))
\end{equation}
where $rank_j(i)$ represents the ranking of item $i$ in the $j$-th individual ranking. Based on the ranking results, we can construct binary tuples containing positive samples and negative samples $(s^+,s^-)$ for the DPO training process.
\begin{figure}[h]
\includegraphics[scale=0.75]{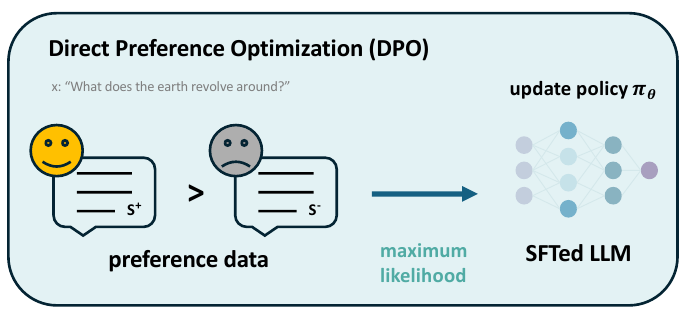}
\centering
\caption{Increasing the log probability of a preferred sample versus decreasing the log probability of a non-preferred sample response, the model's policy tends to select the preferred sample.} \label{fig1}
\end{figure}
\subsubsection{Reinforcement Learning}
Finally, we use the labeled ranking data and DPO to further optimize the model, as shown in Figure 2.
Each training sample in our dataset consists of a conversation between a doctor and a patient, a chosen response, and a rejected response generated by the model.

Given the dataset, we extract the dialogue, chosen response, and rejected response for each sample. 

The DPO loss function is defined as follows:
\begin{equation}
loss = -\log \left( \sigma \left( \beta \cdot \left( (\pi_{\text{chosen}} - \pi_{\text{rejected}}) - \frac{\gamma}{\beta} \right) \right) \right)
\end{equation}
where $\pi_{\text{chosen}}$ is the log probability of the chosen response, and $\pi_{\text{rejected}}$ is the log probability of the rejected response. This ratio measures how much more likely the model is to generate the chosen response compared to the rejected one. The hyperparameter $\gamma$ and $\beta$ controls the adjustment applied to the log probability difference, while scales the entire loss function. The pseudo-code for this phase is shown in Algorithm 2.

\begin{algorithm}[!ht]
    \renewcommand{\algorithmicrequire}{\textbf{Input:}}
	\renewcommand{\algorithmicensure}{\textbf{Output:}}
	\caption{Direct Preference Optimization}
    \label{power}
    \begin{algorithmic}[1] 
        \REQUIRE  SFTed policy $\hat{\pi}_{\theta}$, Traning dataset $\hat{D}$, Number of epochs $N$, Batchsize $B$, Learning rate $\eta\in[0,1]$, sample size $k$, Initial $M=\emptyset$ 
	    \ENSURE Trained policy $\hat{\pi}_\theta$ 
        
        \FOR{$(x,z)\in \hat{D}$}
            \STATE $i=0$
            \WHILE{$i<k$}
                \STATE $M=M\cap \hat{\pi}_{\theta}(x)$
            \ENDWHILE
        \ENDFOR
        \STATE $\widetilde{D}=\text{AutoAnnotation($M$)}$ //Get dpo training data
        \FOR {epoch=1,2,3,...,$N$}
            \FOR{$\widetilde{D}_{batch}\sim \widetilde{D}$}
                \STATE $(s^+,s^-,z)\sim \widetilde{D}_{batch}$
                \STATE $\hat{z}\sim \hat{\pi}_{\theta}(s^+,s^-)$
                \STATE $L = \text{SigmoidLoss}(\hat{z}, z)$
                \STATE $\theta \leftarrow \theta - \eta \nabla_{\theta} L$ //Update the parameters using any gradient descent algorithm
            \ENDFOR
        \ENDFOR

    \end{algorithmic}
\end{algorithm}



\section{Experiments and Evaluation}
\subsection{Training Details}
We use four widely used open-source models as our base models, GLM-4-9B-Chat, Llama-3-8B-instruct\cite{chatglm4}, Qwen2-7B-chat, DeepseekMOE-16B\cite{dai2024deepseekmoe}. GLM-4-9B-Chat is the open-source version of the latest generation of LLms in the GLM-4 family from Smart Spectrum AI, and has been evaluated on a variety of benchmarks in semantics, math, reasoning, code and knowledge extraction, demonstrated superior performance against previous generation of models. LLaMA3 is a LLM developed by Meta, optimized for a wide range of conversational use cases, outperforming many open-source models on common industry benchmarks. Qwen2 is a series of open source large models of Tongyi Qianqian developed by Aliyun. The series provides multiple versions and scales of open source models, such as Base and Instruct, so as to meet different computing needs. DeepSeekMoE 16B is a Mixture-of-Experts (MoE) language model with 16.4B parameters. It employs an innovative MoE architecture, which involves two principal strategies: fine-grained expert segmentation and shared experts isolation. 

Training was performed on 1 NVIDIA Tesla A40(48GB) using a low-rank adaptive (lora) parameter efficient tuning method\cite{hu2021lora}. We trained using bf16 (Deepseek for fp32) precision with learning rate = 5e-5, batchsize = 2, gradient accumulation = 8, maximum length = 1024, dropout = 0.1, and a cosine learning rate scheduler. In the stage of indicating the model produces diversified outputs, we set the number of samples k = 3 and the temperature = 1.2. The statistical information of the dataset is presented in the following Table 3.
\begin{table}[h!]
\centering
\caption{Dataset Composition}
\label{tab:dataset}
\begin{tabular}{|c|c|c|}
\hline
\textbf{Component} & \textbf{Initial Visits} & \textbf{Follow-up Visits} \\
\hline
Training & 50 & 131 \\
\hline
Validation & 7 & 19 \\
\hline
Test & 14 & 38 \\
\hline
\textbf{Total} & \textbf{71} & \textbf{188} \\
\hline
\end{tabular}
\end{table}

\begin{table*}[h!]
\centering
\caption{Performance metrics for various models and methods (Part 1).}
\begin{tabular}{cccccc}
\toprule
 & \multicolumn{5}{c}{Metrics} \\
\cmidrule(lr){2-6}
Model & Method & ROUGE-1 & ROUGE-2 & ROUGE-L & BLEU-4 \\
\midrule
\multirow{4}{*}{GLM4-9b-chat} & Zero-shot & 17.205 $\pm$ 0.083 & 5.403 $\pm$ 0.403 & 11.157 $\pm$ 0.049 & 3.887 $\pm$ 0.142 \\
 & Few-shot & 22.427 $\pm$ 0.046 & 11.064 $\pm$ 0.329 & 18.150 $\pm$ 0.222 & 9.707 $\pm$ 0.509 \\
 & SFT & 54.730 $\pm$ 0.821 & 39.024 $\pm$ 1.320 & 52.453 $\pm$ 0.892 & 37.021 $\pm$ 1.288 \\
 & SFT+DPO & \textbf{55.137} $\pm$ \textbf{0.730} & \textbf{39.718} $\pm$ \textbf{0.392} & \textbf{53.785} $\pm$ \textbf{0.849} & \textbf{38.754} $\pm$ \textbf{0.600} \\
\midrule
\multirow{4}{*}{LLaMA3-8b-chat} & Zero-shot & 4.266 $\pm$ 0.720 & 1.280 $\pm$ 0.090 & 3.076 $\pm$ 0.356 & 1.186 $\pm$ 0.175 \\
 & Few-shot & 18.933 $\pm$ 0.369 & 10.428 $\pm$ 0.386 & 15.707 $\pm$ 0.596 & 7.248 $\pm$ 0.133 \\
 & SFT & \textbf{51.707} $\pm$ \textbf{2.943} & 34.545 $\pm$ 3.401 & 49.269 $\pm$ 2.820 & 31.601 $\pm$ 2.515 \\
 & SFT+DPO & 51.107 $\pm$ 0.426 & \textbf{35.161} $\pm$ \textbf{0.711} & \textbf{50.254} $\pm$ \textbf{0.928} & \textbf{33.934} $\pm$ \textbf{0.627} \\
\midrule
\multirow{4}{*}{DeepSeek-MOE-16b-chat} & Zero-shot & 11.273 $\pm$ 0.689 & 4.107 $\pm$ 0.545 & 10.239 $\pm$ 0.409 & 4.710 $\pm$ 0.296 \\
 & Few-shot & 17.939 $\pm$ 0.223 & 7.496 $\pm$ 0.808 & 13.801 $\pm$ 0.647 & 5.864 $\pm$ 0.763 \\
 & SFT & 51.600 $\pm$ 0.866 & 34.812 $\pm$ 0.723 & 49.787 $\pm$ 0.759 & 32.756 $\pm$ 0.961 \\
 & SFT+DPO & \textbf{54.989} $\pm$ \textbf{1.267} & \textbf{39.576} $\pm$ \textbf{1.473} & \textbf{53.698} $\pm$ \textbf{1.840} & \textbf{37.429} $\pm$ \textbf{1.427} \\
\midrule
\multirow{4}{*}{Qwen2-7b-chat} & Zero-shot & 10.029 $\pm$ 0.291 & 3.873 $\pm$ 0.351 & 11.074 $\pm$ 0.217 & 4.618 $\pm$ 0.577 \\
 & Few-shot & 14.422 $\pm$ 0.291 & 5.213 $\pm$ 0.372 & 11.279 $\pm$ 0.247 & 4.258 $\pm$ 0.278 \\
 & SFT & 51.578 $\pm$ 1.129 & 35.146 $\pm$ 1.748 & 49.309 $\pm$ 1.345 & 33.348 $\pm$ 2.206 \\
 & SFT+DPO & \textbf{53.628} $\pm$ \textbf{1.268} & \textbf{38.163} $\pm$ \textbf{1.649} & \textbf{51.789} $\pm$ \textbf{1.512} & \textbf{36.460} $\pm$ \textbf{1.628} \\
\midrule
\multirow{2}{*}{GPT-3.5-turbo} & Zero-shot & 19.661 $\pm$ 0.584 & 8.865 $\pm$ 0.291 & 16.111 $\pm$ 0.211 & 6.935 $\pm$ 0.343 \\
 & Few-shot & \textbf{32.019} $\pm$ \textbf{1.743} & \textbf{20.040} $\pm$ \textbf{1.784} & \textbf{31.062} $\pm$ \textbf{1.753} & \textbf{18.448} $\pm$ \textbf{1.498} \\
\bottomrule
\end{tabular}
\end{table*}
\subsection{Baseline}
In order to fully evaluate our method, we chose a series of LLMs with different training path and training data as baselines for comparison.
\begin{itemize}
    \item \textbf{Zero-shot\cite{wang2019survey}:} The model is given a task without any prior examples. This approach tests the model's ability to generalize from its training data to new, unseen scenarios. For each model, the zero-shot prompt is carefully crafted to be clear and concise, providing only the necessary context and the query.
    \item \textbf{Few-shot\cite{wang2020generalizing}:} The model is provided with a few examples to learn from before it is asked to perform the task. This method helps the model understand the task better by seeing similar instances. The few-shot prompts are designed to include a small number of examples (2 in our experiment) before presenting the actual query.
    \item \textbf{SFT\cite{brown2020language}:} Supervised fine-tuning will provide medical cases with their corresponding standard diagnoses for the model to learn.
    \item \textbf{SFT+DPO:} After warming up using supervised fine-tuning, the model is guided to generate multiple outputs, and the data is labeled using an automated labeling system which in turn generates preference data for the dpo training process.
\end{itemize}

Zero-shot learning is an important method for evaluating the generalization ability of language models without any task-specific data. For the field of TCM, zero-shot testing can demonstrate the adaptability of the model to unseen TCM tasks, especially when data is scarce. This method can highlight the basic language understanding and reasoning ability of the model, and provide a lower bound on the performance of tasks in this field.

Few-shot learning provides the ability of the model to learn quickly from a small number of examples. Through training with a few samples, we can evaluate whether the model can effectively reason and generate accurate results under the condition of limited labeled data.

SFT is a standard method for model fine-tuning. For the field of TCM, supervised fine-tuning of the model using existing medical case data can improve the performance of the model on specific tasks. In the baseline comparison, supervised fine-tuning provides the upper limit of the performance that the model can achieve after using domain-specific data.

The SFT+DPO method combines supervised fine-tuning and preference optimization to improve the quality of model generation while reducing bias. By using an automated annotation system to generate preference data, DPO can help the model learn outputs that are more in line with the doctor's preferences.
\subsection{Evaluation Metrics}
The assessment of the quality of medical dialog is a multifaceted task. In order to comprehensively assess the quality of medical dialog, we used Bleu\cite{papineni2002bleu}, Rough\cite{lin2004rouge}, and bert-score\cite{zhang2019bertscore} as evaluation metrics. For the same problem, we sampled three outputs from the model and calculated the average metric of these three outputs for error reduction and their standard deviation.

\subsection{Results}
As shown in Tables 4 and 5, the experimental results comprehensively evaluate the performance metrics of various models and methods applied to the proposed TCM task. The primary performance metrics include ROUGE-1, ROUGE-2, ROUGE-L, BLEU-4, precision, recall, and BERT-Score F1. The zero-shot results demonstrate the initial capabilities of the models, generating responses based solely on pre-trained knowledge without any prior examples. In this scenario, all models performed poorly as they had not learned how to prescribe correctly based on medical cases during the pre-training phase. Notably, LLaMA3 performed significantly worse than other models, possibly due to LLaMA being more extensively trained on English corpora, while TCM tasks likely require a higher level of proficiency in Chinese.

After adopting the few-shot method, the performance of all models improved to varying degrees, highlighting their ability to learn and adapt from a small number of instances, thereby enhancing their response quality. It is worth noting that GPT-3.5-turbo showed the most significant improvement, indicating its strong ability to follow instructions. The SFT stage embedded more specific TCM knowledge and prescription tasks into the models, enabling them to generate more accurate and contextually appropriate responses, significantly improving performance across all models at this stage. The DPO stage refined the model outputs using preference data, aligning them more closely with human preferences and expectations, achieving superior overall performance. For certain models, such as LLaMA3, the bias in outputs was reduced post-DPO, indicating the effectiveness of aligning the work with preference data, leading to more stable and intention-aligned outputs.

Despite GPT-3.5-turbo's excellent performance, models enhanced through SFT and DPO exhibited superior results, underscoring the effectiveness of our proposed framework. Considering the difficulty of obtaining high-quality TCM data, achieving such results with a small amount of data is gratifying. Additionally, our framework can be applied to any large foundational model without any modifications. Our research results validate the effectiveness of the proposed framework, particularly the integration of SFT and DPO in improving model performance. The models can generate accurate and relevant medical responses while ensuring alignment with expert preferences, laying the foundation for our next steps in research.

\subsection{Case Study}
To illustrate the effectiveness of our proposed framework, we conducted a detailed case study focusing on its application in Traditional Chinese Medicine (TCM) diagnosis and prescription tasks. This case study demonstrates the model's capability to handle both initial and follow-up diagnoses effectively, showcasing its practical utility in real-world medical scenarios.
We selected a patient case from our corpus, which includes both initial and follow-up visits. The case details are as follows:\par
\subsubsection{Initial Visit}
November 6, 2022.\par

Patient: Liu, Female, 71 years old.\par
Chief Complaint: Dizziness for over a year.\par
Present Illness History: The patient started experiencing dizziness a year ago, occurring when lying flat and turning her head to the left at night, as well as when looking upward during the day. This dizziness is accompanied by visual rotation, occasional nausea and vomiting, a feeling of heaviness and fatigue in the limbs, irritability, normal appetite and sleep, a slightly bitter taste in the mouth, and regular bowel movements and urination.\par
Tongue: Pale tongue with yellow greasy coating. Pulse: Slippery.\par
\subsubsection{Follow-up Visit}

November 13, 2022.\par
Symptoms: After taking the previous prescription, the patient's dizziness significantly reduced. She experienced slight dizziness when getting up in the morning but had almost no dizziness when turning her head to the left. The patient felt physically strong, not irritable, had a normal appetite and sleep, a neutral taste in her mouth, and regular bowel movements and urination. Her tongue was red with a yellow greasy coating, and her pulse was wiry, tight, and rapid.\par
\begin{table*}[h!]
\centering
\caption{Performance metrics for various models and methods (Part 2).}
\begin{tabular}{ccccc}
\toprule
 & \multicolumn{4}{c}{Metrics} \\
\cmidrule(lr){2-5}
Model & Method & Precision & Recall & F1 \\
\midrule
\multirow{4}{*}{GLM4-9b-chat} & Zero-shot & 0.604 $\pm$ 0.001 & 0.748 $\pm$ 0.001 & 0.668 $\pm$ 0.001 \\
 & Few-shot & 0.637 $\pm$ 0.002 & 0.780 $\pm$ 0.003 & 0.700 $\pm$ 0.002 \\
 & SFT & 0.821 $\pm$ 0.003 & 0.830 $\pm$ 0.002 & 0.829 $\pm$ 0.004 \\
 & SFT+DPO & \textbf{0.833} $\pm$ \textbf{0.006} & \textbf{0.834} $\pm$ \textbf{0.003} & \textbf{0.830} $\pm$ \textbf{0.004} \\
\midrule
\multirow{4}{*}{LLaMA3-8b-chat} & Zero-shot & 0.491 $\pm$ 0.002 & 0.544 $\pm$ 0.005 & 0.515 $\pm$ 0.003 \\
 & Few-shot & 0.602 $\pm$ 0.004 & 0.748 $\pm$ 0.014 & 0.666 $\pm$ 0.006 \\
 & SFT & \textbf{0.829} $\pm$ \textbf{0.011} & 0.819 $\pm$ 0.010 & 0.821 $\pm$ 0.010 \\
 & SFT+DPO & 0.824 $\pm$ 0.003 & \textbf{0.827} $\pm$ 0.003 & \textbf{0.823} $\pm$ \textbf{0.003} \\
\midrule
\multirow{4}{*}{DeepSeek-MOE-16b-chat} & Zero-shot & 0.623 $\pm$ 0.002 & 0.745 $\pm$ 0.002 & 0.677 $\pm$ 0.001 \\
 & Few-shot & 0.586 $\pm$ 0.005 & 0.670 $\pm$ 0.005 & 0.622 $\pm$ 0.005 \\
 & SFT & 0.821 $\pm$ 0.006 & 0.825 $\pm$ 0.006 & 0.821 $\pm$ 0.006 \\
 & SFT+DPO & \textbf{0.845} $\pm$ \textbf{0.009} & \textbf{0.834} $\pm$ \textbf{0.010} & \textbf{0.831} $\pm$ \textbf{0.009} \\
\midrule
\multirow{4}{*}{Qwen2-7b-chat} & Zero-shot & 0.601 $\pm$ 0.002 & 0.735 $\pm$ 0.004 & 0.660 $\pm$ 0.002 \\
 & Few-shot & 0.581 $\pm$ 0.003 & 0.701 $\pm$ 0.006 & 0.634 $\pm$ 0.004 \\
 & SFT & 0.822 $\pm$ 0.006 & 0.821 $\pm$ 0.005 & 0.818 $\pm$ 0.005 \\
 & SFT+DPO & \textbf{0.826} $\pm$ \textbf{0.004} & \textbf{0.830} $\pm$ \textbf{0.003} & \textbf{0.826} $\pm$ \textbf{0.003} \\
\midrule
\multirow{2}{*}{GPT-3.5-turbo} & Zero-shot & 0.634 $\pm$ 0.003 & 0.765 $\pm$ 0.001 & 0.692 $\pm$ 0.002 \\
 & Few-shot & \textbf{0.707} $\pm$ \textbf{0.008} & \textbf{0.800} $\pm$ \textbf{0.003} & \textbf{0.748} $\pm$ \textbf{0.006} \\
\bottomrule
\end{tabular}
\end{table*}
\subsubsection{Model application}
We instructed the model to make a corresponding diagnosis based on the patient's symptoms.
\begin{itemize}
    \item \textbf{Step 1: Initial Diagnosis and Prescription}
\end{itemize}

During the initial diagnosis, the model generated the following prescription based on the given patient information:

Traditional Chinese Medicine Diagnosis: Vertigo, hyperactivity of liver yang, and phlegm-dampness obstruction in the middle burner. 

Treatment Principle: Pacify the liver and subdue yang, transform phlegm and dampness. 

Prescription: Modified Tianma Gouteng Decoction combined with Yidou Powder.

Ingredients: Gastrodia (Tianma) 10g, Uncaria (Gouteng) 15g, Raw Dragon Bone (Sheng Longgu) 30g (decoct first), Chuanxiong (Ligusticum Chuanxiong) 10g, Sparganium (Sanleng) 10g, Curcuma Zedoaria (Ezhu) 10g, Angelica Sinensis (Danggui) 15g, Raw Astragalus (Sheng Huangqi) 30g, Leonurus (Yimucao) 30g, White Peony (Baishao) 20g, Ziziphus Jujuba Seed (Zao Ren) 20g, Poria (Fuling) 15g, Polygala (Yuanzhi) 10g, Acorus Tatarinowii (Shichangpu) 10g, Prepared Pinellia (Qing Banxia) 10g, Dried Tangerine Peel (Chenpi) 10g, Bamboo Shavings (Zhuru) 10g, Honey-fried Licorice (Zhigancao) 10g, Amomum (Doukou) 10g, Raw Coix Seed (Sheng Yiyiren) 30g. 

Dosage: 7 doses, to be decocted in water and taken once daily, divided into morning and evening doses.\par
\begin{itemize}
    \item \textbf{Step 2: Follow-up Diagnosis and Prescription}
\end{itemize}

For the follow-up visit, the model synthesized the previous prescriptions and patient feedback to generate a new prescription:

Follow-up Prescription: The previous prescription is continued with the addition of Chrysanthemum (Juhua) 10g, for another 7 doses.

We evaluated the generated prescriptions using automated annotation and confirmed the relevance and accuracy of the prescriptions based on traditional medical principles. The model's ability to adapt to patient feedback and effectively adjust treatment plans demonstrates its potential for practical medical applications. We are well aware that manual evaluation will provide a better insight of how the proposed framework performed in real-world application, but due to issues such as time and cost, we use automated evaluation instead of manual evaluation in this work.

\subsubsection{Insights and Impact}
\begin{itemize}
    \item \textbf{Accuracy}: The model accurately diagnosed and prescribed treatment based on TCM principles, reflecting a deep understanding of the medical corpus it was trained on.
\item \textbf{Adaptability}: The model effectively handled follow-up visits, adjusting prescriptions based on patient feedback and progress, similar to a human practitioner.
\item \textbf{Efficiency}: Automated annotations significantly reduced the need for manual labeling, streamlining the process of preference data collection and enhancing model performance.
\end{itemize}

By integrating supervised fine-tuning with automated annotation and direct preference optimization, our framework not only improves model performance but also ensures that the generated outputs align closely with expert knowledge and user preferences. This approach offers a scalable and efficient solution for enhancing large language models in specialized domains like Traditional Chinese Medicine.

\section{Conclusion and Limitations}
In this paper, we propose a framework that combines supervised fine-tuning and direct preference optimization to improve the performance of large language models for Traditional Chinese Medicine tasks. Our proposed approach addresses the unique challenges faced by TCM, such as the scarcity of high-quality data and the expertise required for accurate medical applications.
By utilizing a small but high-quality corpus of TCM and incorporating an automated annotation process, we are able to significantly improve the model's ability to generate accurate and relevant medical prescriptions. Experimental results show that our framework outperforms existing models, including widely used LLMs such as GPT-3.5-turbo, on various evaluation metrics such as ROUGE, BLEU, and BERT-Score.

Our case study further illustrates the practical applicability of the framework in real TCM consultation scenarios, demonstrating the model's ability to effectively handle both initial and follow-up consultations. Automatic annotation proved to be efficient, reducing the need for manual annotation while maintaining high accuracy of the model output.

Despite the promising results, there are limitations to our approach. The reliance on small datasets, while demonstrating the efficiency of the framework, also highlights the potential advantages of larger and more diverse corpora. In addition, our task is limited to the TCM prescription task only, and the expert annotation is not as high quality as the manual annotation.
In conclusion, our framework provides a scalable and efficient solution for augmenting large language models in specialized fields such as TCM, paving the way for future research and development in integrating AI with TCM practices. Scaling up the size of the dataset, introducing expert labeling and developing new tasks are our future work. In addition, high-quality datasets are difficult to obtain, and due to the particularity of traditional Chinese medicine, datasets are almost all Chinese corpus. Appropriate addition of English corpus for mixed training can fully tap the potential of the model.

\section*{Acknowledgment}
The authors thank anonymous reviewers for their insightful
comments. This research was partially supported by grants from the National Natural Science Foundation of China(No.61877051).

\end{document}